\documentclass[manuscript,screen,authorversion]{acmart}
\usepackage{subcaption}
\usepackage{algorithm}
\usepackage{algorithmic}
\usepackage{multirow}
\AtBeginDocument{%
  }

\setcopyright{acmlicensed}
\copyrightyear{2018}
\acmYear{2018}
\acmDOI{XXXXXXX.XXXXXXX}
\acmConference[Conference acronym 'XX]{Make sure to enter the correct
  conference title from your rights confirmation email}{June 03--05,
  2018}{Woodstock, NY}
\acmISBN{978-1-4503-XXXX-X/2018/06}

\makeatletter
\renewcommand{\@fnsymbol}[1]{%
  \ifcase#1\or%
   \textdagger\or
   \textdaggerdbl\or
   \textasteriskcentered\or
   \textsection\or
   \textparagraph\or
   \|\or
   **\or
   \dagger\dagger\or
   \ddagger\ddagger\or
   \else\@ctrerr\fi%
}
\makeatother
\begin{document}

\title{Efficient Video-to-Audio Generation via Multiple Foundation Models Mapper}



\author{Gehui Chen}
\authornote{Equal contribution.}
\email{23125199@bjtu.edu.cn}
\affiliation{%
  \institution{School of Computer Science and Technology, Beijing Jiaotong University}
  \city{Beijing}
  \country{China}
}

\author{Guan'an Wang}
\authornotemark[1]
\authornote{Project leader.}
\email{guan.wang0706@gmail.com}
\affiliation{%
}

\author{Xiaowen Huang}
\authornote{Corresponding author.}
\email{xwhuang@bjtu.edu.cn}

\author{Jitao Sang}
\email{jtsang@bjtu.edu.cn}
\affiliation{%
  \institution{School of Computer Science and Technology, Beijing Jiaotong University}
  \city{Beijing}
  \country{China}
}
\affiliation{%
  \institution{Beijing Key Laboratory of Traffic Data Mining and Embodied Intelligence}
  \city{Beijing}
  \country{China}
}
\affiliation{%
  \institution{Key Laboratory of Big Data \& Artificial Intelligence in Transportation, Ministry of Education}
  \city{Beijing}
  \country{China}
}









\renewcommand{\shortauthors}{G.Chen, G.Wang, X.Huang et al.}

\begin{abstract}
Recent Video-to-Audio (V2A) generation relies on extracting semantic and temporal features from video to condition generative models. Training these models from scratch is resource intensive.
Consequently, leveraging foundation models (FMs) has gained traction due to their cross-modal knowledge transfer and generalization capabilities. One prior work has explored fine-tuning a lightweight mapper network to connect a pre-trained visual encoder with a text-to-audio generation model for V2A.
Inspired by this, we introduce the Multiple Foundation Model Mapper (MFM-Mapper). 
Compared to the previous mapper approach, MFM-Mapper benefits from richer semantic and temporal information by fusing features from dual visual encoders. 
Furthermore, by replacing a linear mapper with GPT-2, MFM-Mapper improves feature alignment, drawing parallels between cross-modal features mapping and autoregressive translation tasks.
Our MFM-Mapper exhibits remarkable training efficiency. It achieves better performance in semantic and temporal consistency with fewer training consuming, requiring only 16\% of the training scale compared to previous mapper-based work, yet achieves competitive performance with models trained on a much larger scale. 
\end{abstract}




\maketitle

\section{Introduction}
As video generation models such as Sora \cite{sora} and MovieGen \cite{moviegen} become increasingly popular, the video-to-audio (V2A) generation has received significant research attention. 
The researchers aim to generate audio for silent video input, ensuring semantic and temporal consistency with visual information for a harmonious viewing experience. 

Most V2A methods typically employ a two-stage approach. 
The first stage involves training specialized models or leveraging certain visual foundation models to extract semantically and temporally informative features from video frame inputs. 
For example, Diff-Foley \cite{difffoley} utilizes visual-audio contrastive learning to train a visual encoder, while TiVA \cite{tiva} trains an autoregressive model to predict low-resolution Mel-spectrogram layouts. 
The second stage focuses on training audio generation models, which can be diffusion models \cite{ldm, ddpm, ddim}, flow matching models \cite{flowmatching, recitedflowmatching}, or autoregressive models \cite{gpt2, transformer}. 
These models are trained to generate semantically and temporally consistent audio conditioned on the features extracted in the first stage. 

However, training V2A models from scratch is data-intensive and costly. This reliance on large-scale pre-training for cross-modal alignment is a common challenge, as models are expected to learn inter-modal relationships from massive datasets.
To mitigate this challenge, some researchers have explored a data-efficient strategy of integrating pre-trained Foundation Models (FMs) from different modalities.  
By leveraging FMs' robust learning and generalization, efficient cross-modal alignment is achieved by fine-tuning lightweight connectors. This approach has demonstrated significant data and computational efficiency in vision-text tasks, while maintaining strong cross-modal understanding and generation abilities \cite{minigpt, minigptv2, timechat, blip2, instructblip, videochat}.

Following the strategy of connecting FMs, V2A-Mapper \cite{v2amapper} pioneered the application of this approach in the V2A field by training a lightweight mapper network to bridge the visual encoder CLIP \cite{clip} and the text-to-audio diffusion model AudioLDM \cite{audioldm}.
However, we believe that there is still room for improvement in feature extraction and cross-modal alignment. 
Firstly, V2A-Mapper directly inputs CLIP frame features into the mapper. 
This lacks effective modeling and extraction of temporal information from the video, resulting in insufficient temporal context for downstream processing. 
Secondly, the mapper network in V2A-Mapper lacks the pre-trained knowledge and generalization capabilities that are often crucial for effectively handling complex cross-modal feature mapping tasks. 
Finally, as the audio generation model, AudioLDM conditions CLAP \cite{clap} features that focus on audio semantics rather than temporal structure. This inherently restricts AudioLDM's ability to generate temporally coherent audio.

We propose MFM-Mapper to overcome limitations in the existing mapper-based method. It employs a more robust and nuanced approach in feature extraction and cross-modal mapping. 
To achieve rich visual representations, MFM-Mapper integrates features from two distinct pre-trained visual encoders. 
This innovative fusion strategy leverages the complementary strengths of diverse visual models, enabling the system to effectively capture both detailed image semantics and crucial video temporal dynamics. 
This results in a comprehensive and informative video representation that serves as a strong foundation for subsequent audio generation. 
For the audio generation stage, MFM-Mapper utilizes AudioLDM-2 \cite{audioldm2}, a model conditioned on pooled AudioMAE \cite{audiomae} features. 
AudioMAE is a masked autoencoder trained to reconstruct masked Mel-spectrogram patches.
Its embeddings inherently encapsulate both temporal and semantic information within the audio domain \cite{audiomae}. 
This makes it an ideal conditional input for high-fidelity V2A synthesis. 
Finally, to bridge visual and audio modalities, MFM-Mapper fine-tunes a GPT-2 \cite{gpt2} model as an autoregressive mapper. 
Leveraging GPT-2's powerful pre-trained capabilities in generalization and sequence modeling, MFM-Mapper achieves effective and adaptable cross-modal feature alignment, enabling seamless translation from rich visual representations to corresponding audio features.
Remarkably, MFM-Mapper outperforms V2A-Mapper and achieves performance comparable to resource-intensive large-scale models with a significantly reduced training scale (16\%), demonstrating notable data efficiency and cost savings.
\newline
Our contributions are listed following:
\begin{itemize}
  \item \textbf{Temporally Aligned Fusion of Visual FMs:} A key innovation of MFM-Mapper is its temporally aligned fusion strategy. By combining multiple visual FMs, we provide the mapper with unprecedentedly rich semantic and temporal information, directly contributing to temporally synchronized and high-quality audio.
  \item \textbf{Leveraging Pre-trained GPT-2 for Cross-Modal Mapping:} We uniquely leverage the power of pre-trained GPT-2 by fine-tuning it as a cross-modal mapper between vision and audio FMs. This strategic choice allows us to capitalize on GPT-2's vast knowledge and generation capabilities.
  \item \textbf{Data-Efficient and Competitive Performance:} MFM-Mapper, trained with just 16\% of the original scale, surpasses V2A-Mapper in semantic and temporal consistency, achieving overall performance comparable to large-scale models. This underscores MFM-Mapper's data efficiency and competitive edge.
\end{itemize}
\begin{figure*}
  \centering
  \includegraphics[width=1.0\textwidth]{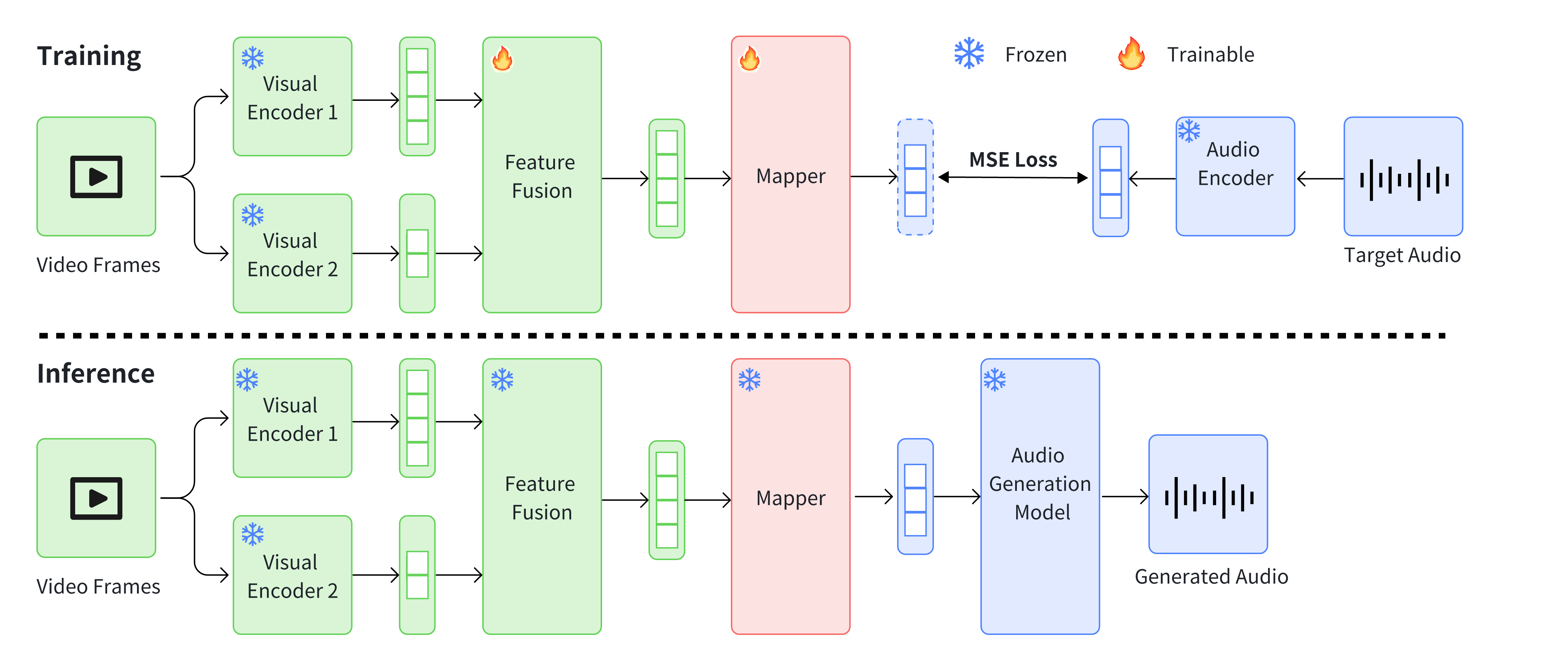}
  \caption{Overall framework of MFM-Mapper. Input video frames are processed by two frozen visual FMs (CAVP and TimeChat). The lower-frequency TimeChat embeddings are replicated to match the higher-frequency CAVP embeddings for temporal alignment. These aligned embeddings are fused and fed into a trainable mapper (fine-tuned GPT-2) to predict pooled AudioMAE embeddings, which then condition the frozen AudioLDM-2 to generate audio. Only the mapper is trained, minimizing the MSE loss between predicted and ground-truth pooled AudioMAE embeddings.}
  \label{fig:frame}
\end{figure*}

\section{Related Works}
\label{sec:relate}
\subsection{V2A generation}
Recent advancements in V2A research \cite{v2amapper, difffoley, mmaudio, seeingandhearing, tiva, vaura, frieren, vatt, foleycrafter, videofoley, condfoley, stav2a} have witnessed a significant surge, actively exploring more advanced architectures and training strategies.
Diffusion \cite{ldm, ddpm, ddim} models and flow matching models \cite{flowmatching, recitedflowmatching} have become mainstream methodologies in the V2A domain.
For instance, Diff-Foley \cite{difffoley} employs contrastive learning on video and audio features to learn cross-modal aligned representations. These representations are then used as conditions to train a latent diffusion model \cite{ldm}
TiVA \cite{tiva} innovatively utilizes low-resolution Mel spectrograms as audio structural information. It trains the model to predict these spectrograms and subsequently conditions a latent diffusion model on them
Frieren \cite{frieren}, based on rectified flow matching \cite{recitedflowmatching}  and incorporating distillation techniques to enhance model efficiency, has demonstrated competitive performance.
MMAudio \cite{mmaudio} integrates video, audio, and text into a unified DiT \cite{dit} network based on flow matching \cite{flowmatching}. By jointly training on audio-visual and text-audio data \cite{vggsound}, MMAudio learns a shared semantic space, enhancing audio quality and alignment, and achieving state-of-the-art performance in V2A.
Autoregressive models like V-AURA \cite{vaura} have also made significant strides in audio quality and temporal alignment.
Seeing-and-Hearing \cite{seeingandhearing} takes a unique approach by leveraging a pre-trained ImageBind \cite{imagebind} model as a cross-modal aligner. It guides audio generation by optimizing alignment scores during inference, showcasing a novel approach to cross-modal control without requiring additional training.
Despite the remarkable progress achieved by these recent methods, designing more lightweight and efficient models for broader application in resource-constrained environments remains an ongoing research direction.
Our work, MFM-Mapper, is proposed in this context, aiming to further advance V2A generation technology in a lightweight and efficient manner by leveraging the power of foundation models.

\subsection{Leveraging FMs for cross-modal learning.} 
Pre-trained on large datasets, FMs show strong feature representations and generalization, enabling effective transfer to many downstream tasks.
FMs are increasingly being leveraged across various cross-modal tasks, and MLLMs in vision-language exemplify this trend. 
By integrating pre-trained visual encoders like CLIP \cite{clip} with LLMs, and often using lightweight adapters for fine-tuning, these models demonstrate strong cross-modal understanding and generation capabilities \cite{minigpt, minigptv2, timechat, blip2, instructblip, videochat}.
This approach reduces the need for extensive cross-modal datasets. 
The success in vision-language motivates their exploration for audio-visual tasks. However, V2A generation presents greater challenges due to the inherent requirement for strict synchronization and semantic consistency between modalities.

\subsection{Mapper-based V2A.}
V2A-Mapper \cite{v2amapper} represents pioneering work in adapting FMs for V2A tasks, exploring lightweight mapping modules to bridge the visual-audio modality gap.
V2A-Mapper specifically trained a lightweight mapping network to map visual embeddings from CLIP into the audio embedding space of CLAP, a pre-trained audio representation model.
This enabled the utilization of AudioLDM \cite{audioldm}, a powerful text-to-audio diffusion model already pre-trained on CLAP \cite{clap} representations, for V2A synthesis, thereby minimizing the number of trainable parameters and enhancing efficiency.
However, it also exhibits inherent limitations due to its reliance on a lightweight mapper without pre-traing and the direct application of CLIP \cite{clip} visual features, which may limit its capacity to fully capture the intricate cross-modal relationships and temporal dynamics crucial in video and audio.
Further improvements in feature extraction and cross-modal mapping methodologies are essential for future progress in the field of V2A generation.

\section{Method}
\label{sec:method}

\subsection{Overall}
\label{sec:overall}
The V2A generation aims to generate semantically consistent and temporally synchronized audio for silent videos. 
A mapper-based V2A method seeks to achieve this by training a mapper network, which takes visual embeddings extracted from a visual foundation model as input, and outputs conditions for a pre-trained audio generation model.

Specifically, a silent video $\mathbf{V}$ with a duration of $T$ is sampled at frames per second (FPS) of $fr$ into a frame sequence $\mathbf{V_f} \in \mathbb{R}^{T_f \times H \times W \times 3}$, where $T_f = fr \times T$, $H$ and $W$ are the height and width of each frame, and $3$ represents the RGB color channels. $\mathbf{V_f}$ is then processed by a visual encoder $\mathcal{F}$:
\[
\mathbf{V_e} = \mathcal{F}(\mathbf{V_f})
\]
The visual encoder $\mathcal{F}$ maps the frame sequence $\mathbf{V_f}$ to a visual embedding sequence $ \mathbf{V_e} \in \mathbb{R}^{T_e \times D_e} $, where $T_e$ is the sequence length and $D_e$ is the embedding dimension. Subsequently, $\mathbf{V_e}$ is fed into a mapper $\mathcal{M}$:
\[
\mathbf{C} = \mathcal{M}(\mathbf{V_e})
\]
The mapper $\mathcal{M}$ transforms $\mathbf{V_e}$ into a condition embedding sequence $\mathbf{C} \in \mathbb{R}^{T_c \times D_c}$, where $T_c$ is the length and $D_c$ is its dimension. Finally, $\mathbf{C}$ is used by an audio generation model $\mathcal{G}$ to generate audio $\mathbf{X}$:
\[
\mathbf{X} = \mathcal{G}(\mathbf{C})
\]

During training, the visual encoder $\mathcal{F}$ and the audio generation model $\mathcal{G}$ are frozen, only the mapper $\mathcal{M}$ is trained to learn the mapping between visual embeddings $\mathbf{V_e}$ and audio generation conditions $\mathbf{C}$. 
By strategically selecting and connecting FMs $\mathcal{F}$ and $\mathcal{G}$ with a trainable mapper $\mathcal{M}$, we hypothesize that pre-trained visual knowledge can be effectively transferred to the audio generation task.
This enables efficient and lightweight V2A generation while simultaneously ensuring high audio quality and synchronization. 

\subsection{Visual Features Fusion}
\label{sec:fusion}
Input visual features are poised to directly influence the performance of the entire V2A pipeline. 
We hypothesize that visual features enriched with effectively extracted semantic and temporal information will establish a robust foundation for subsequent mapper training and ultimately lead to superior audio outputs.
However, relying on a single visual foundation model might prove insufficient to comprehensively capture the diverse aspects of visual information.
Consequently, we propose fusing features derived from pre-trained visual encoders specialized in distinct visual tasks to construct more holistic and high-quality visual features. 
\subsubsection{Visual FMs Selection}
To enhance the semantic relevance and temporal alignment of generated audio, we aim to integrate visual features that effectively capture either semantic or temporal information within video content. Based on this objective, we ultimately selected two visual encoders.

The first encoder is CAVP\cite{difffoley}, which is a video encoder pre-trained by video-audio contrastive learning, derived from Diff-Foley \cite{difffoley}. This approach aims to learn visual representations strongly aligned with synchronized video-audio pair both semantically and temporally. 
Employing a CLIP-like framework \cite{clip}, CAVP uses semantic contrast to maximize audio-visual similarity within videos and minimize it between videos. It also incorporates temporal contrast, emphasizing audio-visual synchronization within video segments. 
This dual contrastive approach enables CAVP to learn rich audio-visual correspondences, providing effective conditioning cues for audio generation, as demonstrated by Diff-Foley's experiments.

The second one is TimeChat\cite{timechat}, which is a cutting-edge video understanding work that combines Visual FMs and LLMs to create a video understanding and time grounding MLLM. 
A key innovation in TimeChat is its cascaded Q-Former \cite{instructblip, blip2} architecture for temporal modeling. This architecture, acting as an intermediate layer, is composed of a time-aware Frame Q-Former and a Sliding Window Q-Former \cite{timechat}.
The Frame Q-Former extracts frame-level visual features from ViT \cite{vit} embeddings, associating them with frame timestamps for temporal awareness.
Subsequently, the Sliding Window Q-Former captures local context within temporal windows and generates a group of embeddings for each window. 
TimeChat's temporal modeling module provides rich visual semantics, precise temporal localization, and effective modeling of continuous inter-frame relationships.

In summary, CAVP focuses on capturing cross-modal audio-visual synchronization associations, while TimeChat's temporal module emphasizes extracting the semantic content of the video and its temporal relationships. The combination of these two is expected to lay a solid foundation for generating semantically relevant and temporally aligned audio.

\subsubsection{Time-Aligned Features Fusion}
Following Section \ref{sec:overall}, we define $\mathcal{F}_1$ and $\mathcal{F}_2$ as the CAVP and TimeChat encoders, respectively.
These encoders process the visual frame sequence $\mathbf{V_f}$ to extract distinct embedding sequences:
\[
\mathbf{V}_{e}^{(1)} = \mathcal{F}_1(\mathbf{V_f}), \quad \mathbf{V}_{e}^{(2)} = \mathcal{F}_2(\mathbf{V_f})
\]
The resulting CAVP embedding sequence $\mathbf{V}_{e}^{(1)}$ belongs to $\mathbb{R}^{{T}_{e}^{(1)} \times {D}_{e}^{(1)}}$, and the TimeChat embedding sequence $\mathbf{V}_{e}^{(2)}$ belongs to $\mathbb{R}^{{T}_{e}^{(2)} \times {D}_{e}^{(2)}}$, where ${T}_{e}^{(1)}$ and ${T}_{e}^{(2)}$ represent the temporal lengths and ${D}_{e}^{(1)}$ and ${D}_{e}^{(2)}$ are their embedding dimensions respectively.  
Our objective is to fuse these two distinct embedding sequences into a unified feature representation, denoted as $\mathbf{V_e}$.  

A crucial preliminary step to fusion is time alignment, which we achieve by upsampling $\mathbf{V}_{e}^{(2)}$ to match the temporal length of $\mathbf{V}_{e}^{(1)}$.  
Subsequently, we employ channel-wise concatenation followed by linear projection as our primary fusion method \cite{attendfusion}, the details of which are elaborated in Algorithm \ref{algorithm:fusion}. 
\begin{algorithm}[!htb]
	\renewcommand{\algorithmicrequire}{\textbf{Input:}}
	\renewcommand{\algorithmicensure}{\textbf{Output:}}
	\caption{Time-Aligned Features Fusion}
	\label{algorithm:fusion}
	\begin{algorithmic}[1]
		\REQUIRE
			$\mathbf{V}_{e}^{(1)} \in \mathbb{R}^{{T}_{e}^{(1)} \times {D}_{e}^{(1)}}$, $\mathbf{V}_{e}^{(2)} \in \mathbb{R}^{{T}_{e}^{(2)} \times {D}_{e}^{(2)}}$ 
		\ENSURE
			$\mathbf{V}_{e} \in \mathbb{R}^{{T}_{e} \times {D}_{e}}$
        \STATE $\mathbf{V}_{e}^{(2)'} = \text{Upsample}(\mathbf{V}_{e}^{(2)}, \text{target\_length} = T_{e}^{(1)})$  
        \STATE $\mathbf{V}_{concat} = \text{Concatenate}(\mathbf{V}_{e}^{(1)}, \mathbf{V}_{e}^{(2)'}, \text{axis=channel})$
        \STATE $\mathbf{V_e} = \text{LinearProjection}(\mathbf{V}_{concat})$
	\end{algorithmic}
\end{algorithm}

Alternatively, a simple additive fusion approach, where $\mathbf{V_e} = \mathbf{V}_{e}^{(1)} + \mathbf{V}_{e}^{(2)'}$, can also be considered.  
Regardless of the specific fusion technique employed, prioritizing time alignment is crucial. Ensuring temporal consistency preserves the temporal structure of the video and provides explicit supervision \cite{moviegen}.

\subsection{Audio Generation Model Selection}
The audio generation model is the core component, responsible for modeling the conditional distribution $P(\mathbf{X}|\mathbf{C})$ to synthesize audio $\mathbf{X}$ (or audio latent) based on condition $\mathbf{C}$. 
The quality and synchronization of audio output depend heavily on the presence of fine-grained semantic and temporal information within $\mathbf{C}$. 
This is not just a preference, but a fundamental constraint: the model's generation capability is limited by the quality of the conditional distribution it learns in training.
Therefore, since mapper-based methods utilize frozen FMs to avoid large-scale training, selecting an audio generation model well trained in high-quality $\mathbf{C}$ is essential for effective and high-fidelity audio results.

V2A-Mapper \cite{v2amapper} utilizes AudioLDM \cite{audioldm} as its audio generation model, leveraging CLAP \cite{clap} features for conditioning. CLAP features, derived from text-audio contrastive learning, are well-suited for capturing global semantic information, but less ideal for capturing the nuanced temporal dynamics necessary for precise audio-visual alignment in V2A. 
In MFM-Mapper, we transition to AudioLDM-2 \cite{audioldm2} for enhanced performance. AudioLDM-2 is a state-of-the-art text-to-audio diffusion model, incorporating pooled AudioMAE \cite{audiomae} embeddings as a key condition.
AudioMAE is a notable encoder-decoder model extensively pre-trained on audio spectrograms with masked reconstruction objectives.
The embeddings extracted from the AudioMAE encoder demonstrate a strong capability to reconstruct heavily masked spectrograms, indicating robust feature representation.
To balance computational efficiency and performance, AudioLDM-2 applies pooling to the AudioMAE embeddings, reducing the 10-second mel-spectrogram patches from 64 time steps and 8 frequency bins to a low-resolution of 8 time steps and 1 frequency bin \cite{audioldm2}.
This pooled AudioMAE, providing information at a temporal granularity of 0.8 FPS, enables AudioLDM-2, despite being primarily designed for text-to-audio synthesis, to exhibit enhanced suitability for V2A applications.

\subsection{Fine-tuning GPT-2 as Mapper}
After feature fusion and audio generation model selection, the input and output of the mapper are specified.
The input is a fused visual embedding sequence $\mathbf{V_e}$, while the output is a sequence of pooled AudioMAE features $\mathbf{C}$, representing a clear modality gap between vision and audio.
We aim to train the mapper $\mathcal{M}(\cdot)$ to learn the mapping from $\mathbf{V_e}$ to $\mathbf{C}$. This mapping demands effective cross-modal alignment and robust relation modeling.  In addition, we desire computational efficiency and a lightweight model, recognizing the trade-off between mapping capacity and training efficiency.
Inspired by the implementation in AudioLDM-2, which fine-tunes GPT-2 in an autoregressive manner, we draw inspiration from their approach of translating Flan-T5 \cite{flant5} embeddings into pooled AudioMAE features \cite{audioldm2}. 
Therefore, analogous to AudioLDM-2's approach, we propose to fine-tune GPT-2 in a similar autoregressive manner, but with different input $\mathbf{V_e}$ and output $\mathbf{C}$. 

For each training sample, $\mathbf{V_e}$ is extracted according to the procedure detailed in Section \ref{sec:fusion}.
Corresponding ground-truth pooled AudioMAE sequence $\mathbf{C}$ is obtained by processing the ground-truth audio through the AudioMAE model. 
We employ the GPT-2 model as our mapper $\mathcal{M(\cdot)}$, and fine-tune it autoregressively to maximize the likelihood according to the following optimization objective \cite{audioldm2}:

\begin{equation}
  \underset{\theta}{\mathrm{argmax}} \  \mathbb{E}_\mathbf{V_e} \prod_{i=1}^{T_c} P(\mathbf{c_i}|\mathbf{c_{i-1}}, \dots, \mathbf{c_0}, \mathbf{V_e}; \theta)
  \label{eq:1}
\end{equation} 
where $\mathbb{E}_\mathbf{V_e}$ means the expectation over the visual embeddings input $\mathbf{V_e}$, $\mathbf{c_i}$ is the i-th embedding of the audio feature sequence $\mathbf{C}$, and $\theta$ represents the parameters to be updated in $\mathcal{M(\cdot)}$. Instead of the standard discrete token prediction via a softmax layer, we fine-tune GPT-2 so that its final Transformer layer directly outputs continuous-valued AudioMAE embeddings. 

During training, we employ a teacher-forcing approach \cite{audioldm2}. 
At each prediction step, the model receives as input the visual features $\mathbf{V_e}$ and the sequence of ground-truth AudioMAE embeddings, $\mathbf{c_1}, \mathbf{c_2}, \dots, \mathbf{c_i-1}$, instead of using the model's own predictions, $\mathbf{\hat{c}_1}, \mathbf{\hat{c}_1}, \dots, \mathbf{\hat{c}_1}$. 
This strategy allows GPT-2 to focus on learning accurate next-step predictions, conditioned on the preceding ground-truth context, thus streamlining the learning process.

We employ the Mean Squared Error (MSE) as the loss function, calculated between the ground-truth AudioMAE features $\mathbf{C}$ and the corresponding predicted features $\mathbf{\hat{C}}$.
\begin{equation*}
    MSE = \frac{1}{T_c} \sum_{i=1}^{T_c} ||\mathbf{c_i} - \mathbf{\hat{c}_i}||^2
\end{equation*}
minimizing the MSE loss directly maximizes the likelihood objective of Equation \eqref{eq:1}. This forces the predicted features $\mathbf{\hat{C}}$ to align with the ground-truth $\mathbf{C}$, thereby optimizing the conditional probabilities $P(\mathbf{c_i}|\mathbf{c_{i-1}}, \dots, \mathbf{c_0}, \mathbf{V_e}; \theta)$ for each embedding in the sequence.

\section{Experiments}
\label{sec:experiments}
\subsection{Dataset and Implementation Details}
\label{sec:dataset}
\subsubsection{Dataset.}

VisualSound \cite{vaura}, a curated subset of the VGGSound \cite{vggsound} dataset, is used to train our model due to its focus on samples with strong audio-visual coherence. 
VGGSound itself is a large dataset comprising approximately 200,000 ten-second YouTube videos across over 300 categories. 
However, a significant portion of the original VGGSound samples exhibit weak audio-visual coherence, including videos with irrelevant audio, such as non-diegetic \cite{nondiegetic} speech, background music, or extraneous noise.
To address this, VisualSound leverages the ImageBind \cite{imagebind} model to identify and filter out videos with poor audio-visual relevance.
This filtering process results in a final training set of over 77,000 samples, less than half the size of the original VGGSound. 
During data processing, we have identified a small number of silent videos and removed them from the training set.
For fair comparison with prior work, we evaluate our approach on the complete VGGSound test set. \newline
\subsubsection{Implementation details.} We have ensured consistency in all data processing, training, and inference steps by following the methods detailed in the original papers and official code releases. 
To extract visual features, we sample each 10-second video clip \cite{clip} at 4 FPS. 
Clips longer than 10 seconds were truncated to 10 seconds, and clips shorter than 10 seconds were zero-padded to achieve a 10-second length. 
The frame sequence is then directly input to the CAVP \cite{difffoley} model. 
However, for TimeChat \cite{timechat}, the frame sequence is initially processed by a ViT to generate patch-level representations before being fed into the Q-Formers \cite{timechat}. 

For AudioLDM-2 \cite{audioldm2}, we use the open-source base version. 
During inference, the number of denoising steps is set to 100, and we use the DDIM \cite{ddim} scheduler. 
We also use a negative prompt 'low quality,average quality' with classifier free guidance (CFG) \cite{cfg} equals 3.5. 
These settings follow the best practices recommended in the original paper \cite{audioldm2}. 
Notably, AudioLDM-2 is conditioned on two multimodal features: FLAN-T5 \cite{flant5} text embeddings and pooled AudioMAE \cite{audiomae} embeddings. Preliminary experiments have revealed that the absence of FLAN-T5 embeddings significantly impacts audio quality. 
To fully leverage its capabilities, during test-time inference, we also provide audio class labels from the dataset as text prompts, such as 'dog bow-wow'. These labels are encoded into text embeddings by FLAN-T5 and used as the second conditional input to the diffusion model.

For fine-tuning, we use the same configuration as AudioLDM-2, including an embedding dimension of 768 and 12 transformer layers. Our GPT-2 model is initialized with the pre-trained weights released by AudioLDM-2.
Training is performed using the AdamW \cite{adam, adamw} optimizer with a learning rate of 1e-3 for 40 epochs and a global batch size of 48. Data parallelism is employed across 3 NVIDIA A4000 GPUs. The training process takes approximately 18 hours.

\subsection{Evaluation Setups}

For a robust and standardized evaluation of our proposed method in V2A synthesis, we directly adopt the evaluation setup provided by MMAudio \cite{mmaudio}.  
MMAudio \cite{mmaudio} generously release their evaluation code and scripts publicly, which implements a comprehensive framework aligned with established practices in the field. By reusing this publicly available evaluation framework, we ensure a fair and direct comparison with other existing models on the VGGSound dataset. 
It's important to note that while our model is trained on 10-second audio clips, the MMAudio evaluation script uses 8 seconds as a standard benchmark. Therefore, during testing, our generated audio samples are automatically cropped to 8 seconds to align with this evaluation protocol.
\newline
\subsubsection{Metrics.} 
To evaluate the generated audio, MMAudio employs a set of four key metrics, covering distribution matching, audio quality, semantic alignment, and temporal alignment. These metrics are implemented in the publicly available MMAudio evaluation scripts.
\begin{itemize}
    \item \textbf{Distribution Matching:} For Distribution Matching, Fréchet Distance (FD) \cite{fad} and Kullback-Leibler (KL) divergence are computed. FD is calculated using PaSST \cite{passt} (FD\textsubscript{PaSST}), PANNS \cite{panns} (FD\textsubscript{PANNs}), and VGGish \cite{audioset} (FD\textsubscript{VGG}) embeddings, while KL divergence utilizes PANNS (KL\textsubscript{PANNs}) and PaSST (KL\textsubscript{PaSST}) classifiers, following the MMAudio approach.
    \item \textbf{Audio Quality:} Audio Quality is evaluated using the Inception Score (IS), which assesses perceptual quality and diversity via a pre-trained audio classification model (PANNS classifier, as implemented in MMAudio).
    \item \textbf{Semantic Alignment:} Semantic Alignment is measured via the ImageBind score (IB-score) \cite{foleygen}, which quantifies semantic coherence by calculating cosine similarity between ImageBind \cite{imagebind} features.
    \item \textbf{Temporal Alignment:} Temporal Alignment, reflecting audio-visual synchrony, is evaluated using the DeSync score. Predicted by Synchformer \cite{synchformer}, DeSync estimates temporal misalignment in seconds, with lower values indicating better synchrony.
\end{itemize}

To demonstrate MFM-Mapper's training efficiency, we report two key metrics: the training scale (Scale) and the number of trainable parameters (Param.). Training scale, representing the total audio duration used for training, is calculated as the product of the total audio time per epoch and the number of training epochs.

\subsubsection{Baselines.} 
For comparative evaluation, we adopted the same baseline models as those benchmarked in MMAudio \cite{mmaudio}, ensuring alignment with their established evaluation protocol. We directly leveraged the baseline results reported by MMAudio for a consistent comparison. Refer to Table \ref{tab:main} for a comprehensive presentation of these comparative results.

\subsection{Main Results}
\begin{table*}[!t]
\centering
\scriptsize 
\begin{tabular}{l *{10}{c}}
\toprule
\multirow{3}{*}{Method} & \multicolumn{2}{c}{Training-efficiency} & \multicolumn{4}{c}{Distribution matching} & \multicolumn{2}{c}{Audio quality} & \multicolumn{1}{c}{Semantic align.} & \multicolumn{1}{c}{Temporal align.} \\
\cmidrule(lr){2-3} \cmidrule(lr){4-7} \cmidrule(lr){8-9} \cmidrule(lr){10-10} \cmidrule(lr){11-11}
& Scale & Param. & FD\textsubscript{PaSST}$\downarrow$ & FD\textsubscript{PANNs}$\downarrow$ & FD\textsubscript{VGG}$\downarrow$ & KL\textsubscript{PANNs}$\downarrow$ & KL\textsubscript{PaSST}$\downarrow$ & IS$\uparrow$ & IB-score$\uparrow$ & DeSync$\downarrow$  \\
\midrule
V-AURA \cite{vaura}$^{*}$ & - & 695M & 218.50 & 14.80 & 2.88 & 2.42 & 2.07 & 10.08 & 27.64 & 0.654 \\
VATT \cite{vatt}$^{*}$ & - & - & 131.88 & 10.63 & 2.77 & \textbf{1.48} & \textbf{1.41} & 11.90 & 25.00 & 1.195 \\
Frieren \cite{frieren}$^{*}$ & - & 159M & 106.10 & 11.45 & 1.34 & 2.73 & 2.86 & 12.25 & 22.78 & 0.851 \\
FoleyCrafter \cite{foleycrafter}$^*$ & - & 1.22B & 140.09 & 16.24 & 2.51 & 2.30 & 2.23 & \textbf{15.68} & 25.68 & 1.225 \\
V2A-Mapper \cite{v2amapper}$^*$ & 50,000h & 48.83M & 84.57 & 8.40 & 0.84 & 2.69 & 2.56 & 12.47 & 22.58 & 1.225 \\
MMAudio-S-16kHz \cite{mmaudio}$^*$ & $>$100,000h & 157M & \textbf{70.19} & \textbf{5.22} & \textbf{0.79} & 1.65 & 1.59 & 14.44 & \textbf{29.13} & \textbf{0.483} \\
\midrule
MFM-Mapper & 8,000h & 124M & 111.74 & 9.90 & 1.74 & 2.26 & 2.23 & 12.22 & 26.07  & 1.073  \\
MFM-Mapper-M & 20,000h & 124M & 94.26 & 8.49 & 0.95 & 2.23 & 2.16 & 10.58 & 25.68 & 1.077 \\
\bottomrule
\end{tabular}
\caption{Video-to-audio results on the VGGSound test set. Values marked with * are directly quoted from MMAudio \cite{mmaudio}. MFM-Mapper results are shown for two training configurations, both using 40\% of the VGGSound dataset but trained for 40 epochs and 100 epochs respectively.}
\label{tab:main}
\end{table*}

\subsubsection{Quantitative Results}
The quantitative results presented in Table \ref{tab:main} underscore the remarkable training efficiency of MFM-Mapper. 
Notably, our model achieves these results with a training scale of only 8,000 hours (MFM-Mapper) and 20,000 hours (MFM-Mapper-M), representing a mere 16\% and 40\% of the training scale used by V2A-Mapper, respectively. 
This drastic reduction in training data highlights the significant data efficiency of our approach.
Despite this considerably smaller training scale, MFM-Mapper demonstrates substantial performance gains over V2A-Mapper. 
Specifically, MFM-Mapper exhibits a significant improvement in IB-score, increasing by 15.5\%, and a notable reduction in DeSync score by 12.4\% compared to V2A-Mapper. MFM-Mapper also achieves better KL\textsubscript{PANNs} and KL\textsubscript{PaSST} scores.
These improvements can be attributed to the key design choices in MFM-Mapper mentioned in Section \ref{sec:method}.
While MFM-Mapper demonstrates clear advantages in semantic and temporal alignment, its Inception Score (IS) is slightly lower than V2A-Mapper, and there is a noticeable gap in Fréchet Distance (FD) metrics. 
Interestingly, V2A-Mapper achieves the second-best FD scores among the baselines. This might be partially explained by V2A-Mapper's training focusing specifically on aligning CLIP visual features with CLAP audio features, potentially leading to a tighter distribution match in the feature space used for FD calculation. 
In contrast, MFM-Mapper learns semantic and temporal consistency over complex distribution matching in the feature space.

Compared to other models trained on significantly larger datasets, such as MMAudio, the superior performance of MMAudio in all metrics is evident, reflecting the benefits of large-scale end-to-end training. Nevertheless, MFM-Mapper maintains strong competitiveness against other baselines with larger parameter counts and training data requirements. As observed in Table \ref{tab:main}, both MFM-Mapper and MFM-Mapper-M consistently rank within the top 3-5 across IB-score, DeSync, KL and FD metrics, excluding MMAudio. 
This demonstrates that MFM-Mapper achieves a compelling balance between training efficiency and competitive performance, offering a viable and resource-friendly approach to high-quality V2A generation.
When extending the training of MFM-Mapper to 100 epochs (MFM-Mapper-M), we observe a considerable improvement in FD metrics, suggesting that further training can indeed bridge this distribution matching gap. 
However, the temporal and semantic alignment metrics (IB-score and DeSync) show no improvement compared to 40 epochs, indicating that these aspects of performance may plateau earlier. This might suggest that the initial 40 epochs are crucial for learning the fundamental cross-modal alignment and temporal dynamics, while subsequent epochs refine the overall audio quality and distribution matching.

\subsubsection{Qualitative Results}
\begin{figure*}
  \centering
  \includegraphics[width=1.0\textwidth]{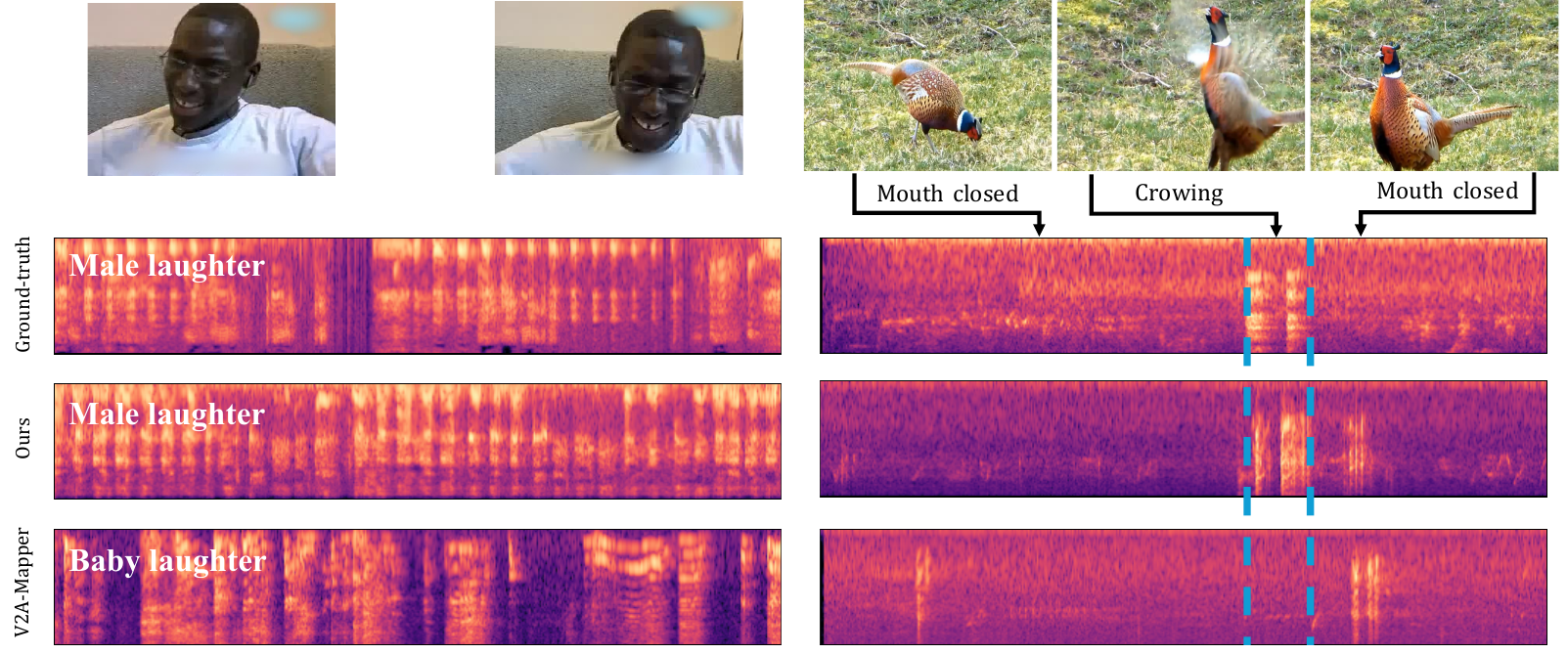}
  \caption{Qualitative comparison of generated audio spectrograms. Each panel displays video frames, the ground-truth spectrogram, and spectrograms generated by MFM-Mapper (Ours) and V2A-Mapper. Left: Example of a man laughing. Right: Example of a pheasant crowing.}
  \label{fig:case}
\end{figure*}

Figure \ref{fig:case} provides a visual comparison of the audio generation quality between our MFM-Mapper, V2A-Mapper, and the ground truth. These examples highlight MFM-Mapper's improved performance in both semantic and temporal consistency. 
In the first case (left panel), concerning male laughter, V2A-Mapper correctly identifies the broad semantic category of laughter but mischaracterizes it as 'Baby laughter'. In contrast, our method accurately generates 'Male laughter,' aligning semantically with the ground truth and the visual input. 
The second case (right panel) demonstrates temporal accuracy: the pheasant's crowing occurs around the 5-second mark, as indicated in the ground-truth spectrogram. MFM-Mapper successfully synchronizes the generated sound event with this timing, whereas V2A-Mapper fails to localize the crowing sound within the correct temporal window indicated by the visual cues.

\subsection{Ablations}

\subsubsection{Multiple visual encoders.} 
\begin{table}[!t]
\centering
\footnotesize 
\begin{tabular}{l *{4}{c}} 
\toprule
\textbf{Visual encoder(s)} & \textbf{FD\textsubscript{PassST}$\downarrow$} & \textbf{IS$\uparrow$} & \textbf{IB-score$\uparrow$} & \textbf{DeSync$\downarrow$} \\
\midrule
CAVP+TimeChat & \textbf{111.74} & \textbf{12.22} & \textbf{26.07} & \textbf{1.073} \\
CAVP & 123.26 & 11.46 & 22.01 & 1.086 \\
TimeChat & 124.24 & 12.08 & 25.52 & 1.162 \\
\bottomrule
\end{tabular}
\caption{Ablation study on visual encoders. Fusing CAVP and TimeChat encoders outperforms using each encoder individually, demonstrating the effectiveness of the proposed fusion strategy.}
\label{tab:ab-multi}
\end{table}
The ablation study in Table \ref{tab:ab-multi} clearly demonstrates the benefits of fusing visual features from both CAVP and TimeChat encoders. 
Across all evaluation metrics, the combined CAVP+TimeChat approach outperforms using either encoder in isolation. This consistent improvement underscores the effectiveness of our proposed visual feature fusion strategy. 
Notably, CAVP, pre-trained with video-audio contrastive learning, is inherently strong in capturing temporal alignment, while TimeChat excels in extracting rich semantic information from video frames. 
This complementary nature is reflected in the individual encoder results: CAVP shows a lower IB-score indicating weaker semantic alignment, and TimeChat exhibits a higher DeSync score suggesting less precise temporal synchronization. 
The successful fusion of these two models effectively leverages their respective strengths, leading to a more comprehensive and robust visual representation beneficial for V2A generation.
\newline
\subsubsection{Time-aligned Fusion}
\begin{table}[!t]
\centering
\footnotesize 
\begin{tabular}{l *{4}{c}} 
\toprule
\textbf{Fusion method} & \textbf{FD\textsubscript{PassST}$\downarrow$} & \textbf{IS$\uparrow$} & \textbf{IB-score$\uparrow$} & \textbf{DeSync$\downarrow$} \\
\midrule
Channel-wise + proj & \textbf{111.74} & 12.22 & \textbf{26.07} & \textbf{1.073} \\
Add & 113.58 & \textbf{12.33} & 26.00 & 1.091 \\
Cat & 112.23 & 12.24 & 24.73 & 1.120 \\
\bottomrule
\end{tabular}
\caption{A comparison of different fusion methods. The
”Channel-wise + proj” and ”Add” methods incorporate time align-
ment, whereas the ”Cat” method does not.}
\label{tab:ab-time}
\end{table}
To demonstrate the importance of fusion with time alignment, we compare different fusion methods. The results are presented in Table \ref{tab:ab-time}. The "Channel-wise + proj" method refers to concatenating the temporally aligned visual feature sequences along the feature dimension followed by a linear layer, the same as the algorithm in \ref{sec:fusion}. 
The "Add" method refers to directly summing the temporally aligned visual embeddings. Both of these methods utilize alignment and maintain the same time sequence length. 
In contrast, the "Cat" method concatenates the visual features along the time dimension, resulting in a longer visual feature sequence. 
The results show that the "Cat" method exhibits a lower performance on Acc compared to the others. This clearly demonstrates the significant impact of temporally aligned fusion on temporal alignment.
\subsubsection{Autoregression vs. Diffusion}
To further evaluate the effectiveness of our chosen autoregressive mapping strategy, we conducted an ablation study comparing it against a diffusion-based mapper, termed "Diff-Mapper". 
For a fair comparison, Diff-Mapper utilizes the same GPT-2 architecture as its backbone network. 
Following the diffusion approach used in previous work like V2A-Mapper \cite{v2amapper} , the GPT-2 network takes as input a noise timestep embedding $t$, the fused visual embeddings $\mathbf{V_e}$ ,the noisy AudioMAE embeddings $\mathbf{C}_t$, and learnable tokens. The final layer's output embeddings corresponding to these learnable tokens serve as the prediction for $\mathbf{C}_{t-1}$.
The model is trained using standard diffusion model objectives.
During inference, this iterative denoising process starts from pure noise $C_T$ and progresses until $\mathbf{C}_0$ is reached, representing the final predicted pooled AudioMAE embeddings $\mathbf{\hat{C}}$.

\begin{figure}[!t]  
  \centering
  \begin{subfigure}[b]{0.48\textwidth}
      \centering
      \includegraphics[width=\textwidth, height=5cm, keepaspectratio]{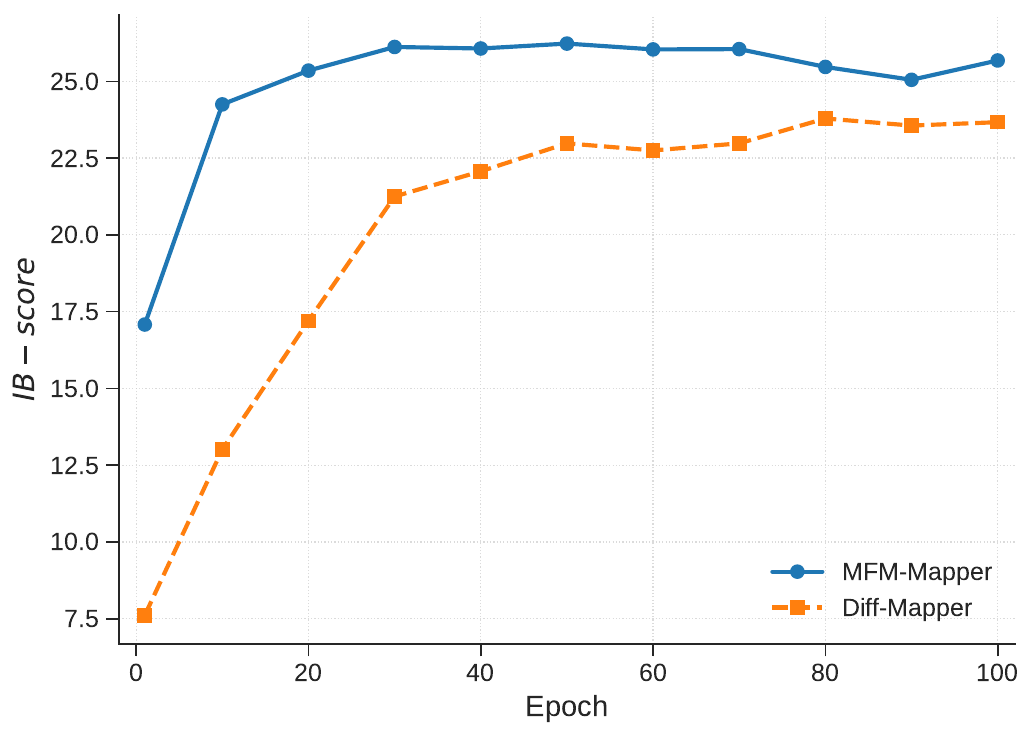}
      \caption{IB-score over Epochs}
      \label{fig:epoch_ib}
  \end{subfigure}
  \hfill
  \begin{subfigure}[b]{0.48\textwidth}
      \centering
      \includegraphics[width=\textwidth, height=5cm, keepaspectratio]{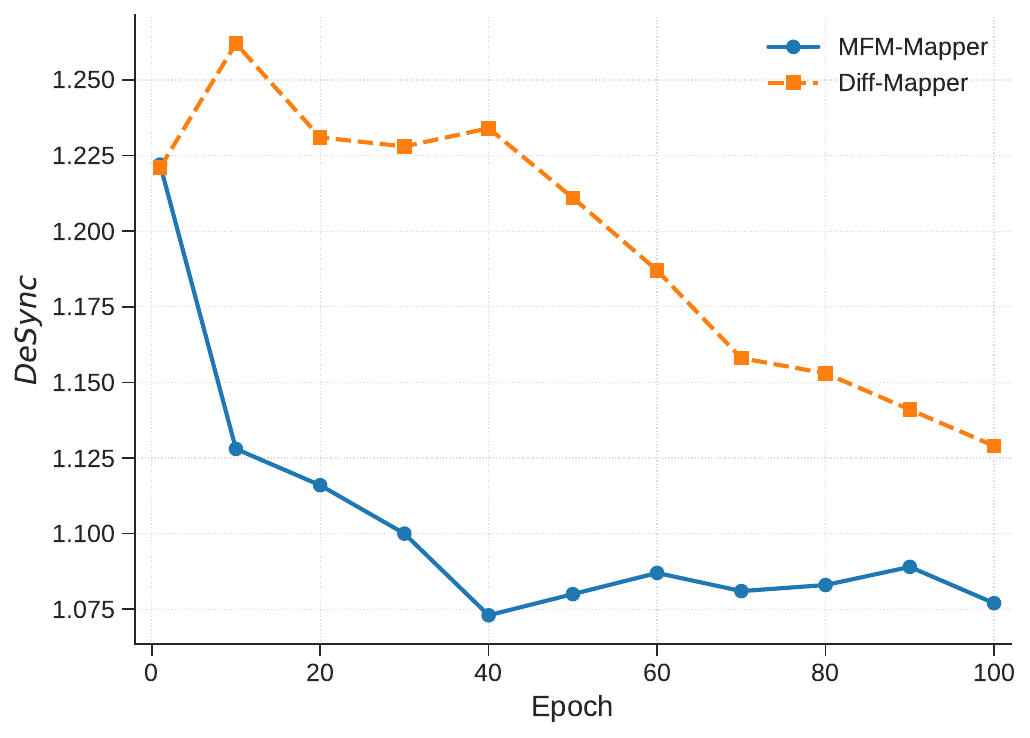}
      \caption{DeSync Score over Epochs}
      \label{fig:epoch_desync}
  \end{subfigure}
  \caption{Training dynamics comparing the MFM-Mapper (Ours) and the Diff-Mapper over 100 epochs on key alignment metrics. (a) Semantic alignment evaluated by IB-score (higher is better). (b) Temporal alignment evaluated by DeSync score (lower is better). Our autoregressive method shows faster convergence and maintains superior performance compared to the diffusion-based alternative.}
  \label{fig:epoch}
\end{figure}
We compare the training dynamics of both approaches by tracking performance over epochs, illustrated in Figure \ref{fig:epoch}. 
Our autoregressive method exhibits significantly faster convergence; most metrics show diminishing returns and approach a plateau around epoch 40. In contrast, the Diff-Mapper typically requires more extensive training to reach its peak performance. 
Crucially, our method consistently outperforms the Diff-Mapper at equivalent training stages across all observed epochs. 
This faster convergence and superior performance profile throughout training strongly underscore the higher training efficiency and effectiveness of our proposed autoregressive mapping strategy compared to the diffusion-based alternative.

\section{Conclusion and Future Work}

This paper introduces MFM-Mapper, a novel approach to V2A generation that leverages multiple FMs and a fine-tuned language model for efficient cross-modal mapping. Our results demonstrate that MFM-Mapper achieves performance comparable to that of state-of-the-art methods in terms of audio quality, semantic relevance, and temporal synchronization, while requiring significantly fewer training resources. The effectiveness of our temporally aligned fusion strategy and utilizing a language FM as a mapper has been clearly shown.

Future work will focus on several key areas. First, we plan to explore more sophisticated fusion techniques to further enhance the temporal alignment and semantic consistency of the generated audio. Second, we will investigate the application of MFM-Mapper to a wider range of video datasets and audio generation tasks to assess its broader generalizability. Finally, exploring the integration of MFM-Mapper with advanced video generation models for seamless audio-visual content creation is a promising direction.

\begin{acks}
This work was supported in part by the National Key Research and Development Program of China under Grant (2023YFC3310700), the National Natural Science Foundation of China (62572040, 62202041), and the Beijing Natural Science Foundation (JQ24019).
\end{acks}

\bibliographystyle{ACM-Reference-Format}
\bibliography{sample-base}










\end{document}